\documentclass[conference]{IEEEtran}
\IEEEoverridecommandlockouts
\usepackage[top=0.7in,bottom=0.95in, left=0.7in, right=0.7in]{geometry}

\usepackage{array}
\usepackage{cite}
\usepackage{amsmath,amssymb,amsfonts}
\usepackage{algorithmic}
\usepackage{algorithm}
\usepackage{graphicx}
\usepackage{textcomp}
\usepackage{xcolor}
\usepackage{hyperref}
\usepackage[numbers]{natbib}
\def\BibTeX{{\rm B\kern-.05em{\sc i\kern-.025em b}\kern-.08em
    T\kern-.1667em\lower.7ex\hbox{E}\kern-.125emX}}

\usepackage{etoolbox}
\makeatletter
\patchcmd{\@makecaption}{\scshape}{}{}{}
\patchcmd{\@makecaption}{\\}{.\ }{}{}
\makeatother
\def\tablename{Table}

\begin{document}

\title{Semi-decentralized Federated Time Series Prediction with Client Availability Budgets}

\author{
\IEEEauthorblockN{Yunkai Bao\IEEEauthorrefmark{1}, Reza Safarzadeh\IEEEauthorrefmark{2}, Xin Wang\IEEEauthorrefmark{2}, Steve Drew\IEEEauthorrefmark{1}}
\IEEEauthorblockA{
\IEEEauthorrefmark{1}Department of Electrical and Software Engineering, University of Calgary, Calgary, AB, Canada \\
\IEEEauthorrefmark{2}Department of Geomatics Engineering, University of Calgary, Calgary, AB, Canada \\
\{yunkai.bao, reza.safarzadeh, xcwang, steve.drew\}@ucalgary.ca}
}

\maketitle

\begin{abstract}

Federated learning (FL) effectively promotes collaborative training among distributed clients with privacy considerations in the Internet of Things (IoT) scenarios. Despite of data heterogeneity, FL clients may also be constrained by limited energy and availability budgets. Therefore, effective selection of clients participating in training is of vital importance for the convergence of the global model and the balance of client contributions.
In this paper, we discuss the performance impact of client availability with time-series data on federated learning. We set up three different scenarios that affect the availability of time-series data and propose FedDeCAB, a novel, semi-decentralized client selection method applying probabilistic rankings of available clients. When a client is disconnected from the server, FedDeCAB allows obtaining partial model parameters from the nearest neighbor clients for joint optimization, improving the performance of offline models and reducing communication overhead.
Experiments based on real-world large-scale taxi and vessel trajectory datasets show that FedDeCAB is effective under highly heterogeneous data distribution, limited communication budget, and dynamic client offline or rejoining.

\end{abstract}


\section{Introduction}
% The private nature of on-device personal data, including photo albums, browsing histories, images and videos captured by connected autonomous vehicles (CAVs), and the activity history from wearable devices, has collectively called for a privacy-aware distributed learning paradigm from distributed devices where federated learning (FL) is a promising solution~\cite{mcmahan2017communication}.
%
% In contrast to centralized machine learning, FL clients train models using their data and only transmit model weights to a server without training data, suitable for edge computing scenarios with varying topologies \cite{wu2023topology}.

Traditional machine learning (ML) requires the central server to gather data from various node devices for model training. However, the access to personal data from edge devices is usually subject to privacy restrictions and the large scale of local data in common edge computing environments or scenarios with a large number of Internet of Things (IoT) devices. Federated learning (FL), as a distributed machine learning paradigm, breaks away from the reliance of traditional ML on centralized data \cite{mcmahan2017communication}. FL users train models locally on their personal data and only transmit their model parameters to the central server without compromising the data privacy.

\begin{figure}[t]
\centering
\includegraphics[scale=0.22]{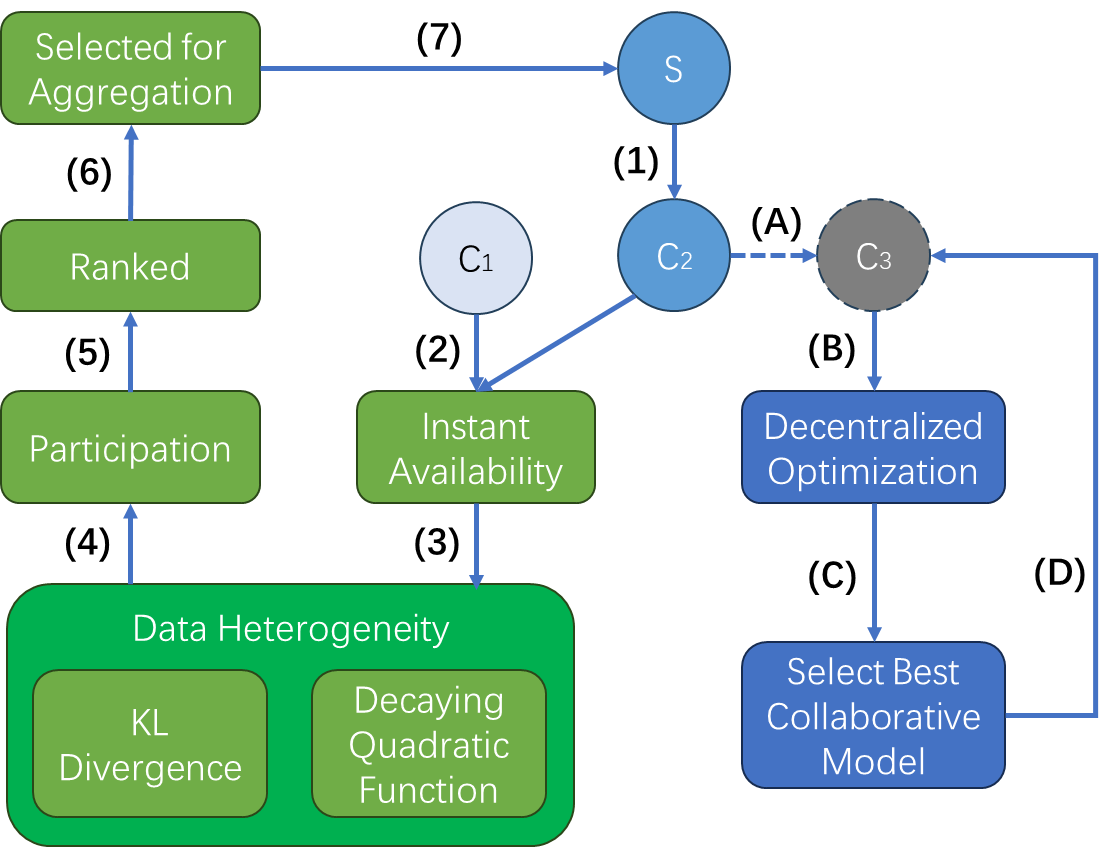}
% \caption{Overview of FedCAB with 7 steps: (1) Download the latest global model from server to client. (2) Only consider instantly available clients for the next round. (3) Use Kullback–Leibler (KL) divergence with a quadratic decay function to evaluate data heterogeneity. (4) Adjust ranking, factoring in client participation. (5) Sort weights in descending order. (6) Select clients by weight. (7) Model aggregation with selected clients. }
\caption{Overview of FedDeCAB with two frameworks: centralized and offline FL. Centralized FL includes 7 steps: (1) Download the latest global model from the server to continuously online clients (C2), while newly online clients (C1) use local model for update. (2) Only consider instantly available clients for the next round. (3) Use Kullback-Leibler (KL) divergence and a quadratic decay function to evaluate data heterogeneity. (4) Adjust ranking, factoring in client participation. (5) Arrange weights in descending order. (6) Select clients by weight. (7) Model aggregation with the selected clients. Offline FL includes 4 steps: (A) Offline clients (C3) obtain parameters from all neighboring clients. (B) Update with local model and the collaborative model from previous round. (C) Calculate empirical loss using all obtained parameters. (D) Selects model with minimum loss as the collaborative model for the next round.}
\label{fig:feddecab-steps}
\end{figure}

One of the bottlenecks in FL comes from the heterogeneous computational and communication capabilities across different edge devices~\cite{diao2021heterofl}, which causes FL clients to participate in the training process with an availability budget, explicitly or implicitly. The availability budgets are affected by the properties of the edge devices and the network communication, including energy-constrained device power supply, unpredictable device mobility, unstable wireless connections, and the diversity of edge network topologies. Lack of availability can prevent devices from participating in the training and updating processes, either temporarily or permanently~\cite{zhu2022resilient}. This phenomenon is called client dropout \cite{horvath2021fjord} or client stragglers \cite{chai2020tifl}, which would hinder the convergence of the global model. In addition, the asynchrony of edge users causes different clients to join FL training at different stages instead of participating all at the same initial round. Clients with heterogeneous data distribution that join FL at a later stage will train on a relatively stable global model, weakening their contributions or disrupting the further convergence of the global model. We refer to these asynchronous clients as late-join clients.

% In order to address these challenges mentioned above, we propose FedCAB in this paper, a novel probabilistic ranking algorithm for client selection in FL, which considers the impact of client availability budgets to achieve faster convergence of the global model.
FedCAB \cite{bao2023fedcab} is a novel probabilistic ranking algorithm for client selection in FL, which considers the impact of client availability budgets to achieve faster convergence of the global model. Numerical experiments were conducted on image classification datasets such as MNIST \cite{cohen2017emnist} and Fashion-MNIST \cite{ribero2022fashionmnist}. However, no experiments have been conducted to demonstrate whether FedCAB can maintain good performance on time-series prediction tasks with data availability constraints and unstable client connectivity.

In the FL time-series prediction task, the system will face the constraint of data availability. Maritime data collected by AIS receivers or vehicle trip data sampled by GPS sensors have long been widely used in trajectory prediction tasks  \cite{Liang2021NetTrajAN}. Trajectory prediction plays a vital role in traffic monitoring, scheduling, and collision avoidance \cite{forti20traj}. It is also widely used in IoT-related applications such as smart cities or autonomous driving \cite{libcity}. In the FL system based on real-time collection of trajectory data by edge clients such as ships or vehicles, the availability of data is affected by the energy-constrained sampling frequency and the connectivity with the positioning system. Therefore, in the real-time trajectory prediction task, the distribution of trajectory data points may be more intermittent and discrete. Existing works on compensating for missing trajectory points infer missing values or generate similar trajectories \cite{xia23trajrec} \cite{yu22trajrec} \cite{nguyen18recons}, but these methods tend to introduce additional computational burdens or privacy concerning on edge clients.

The unavailability of edge clients in the FL time-series prediction may also lead to severe communication straggling, manifested by multiple disconnections from the central server and re-connection after a period of time, rather than just delayed joining like late-join clients. We refer to clients disconnected from the central server as offline clients, and clients that are able to communicate with the server as online clients. However, it is still possible for an offline client to maintain connections with its neighboring clients, making it possible to establish a semi-decentralized collaborative optimization approach.

In order to address these challenges mentioned above, we propose FedDeCAB in this paper. FedDeCAB retains the probabilistic ranking algorithm proposed in FedCAB, as well as the consideration of client availability budgets. In addition, we extend this work into a more complicated communication environment and introduce a semi-decentralized mechanism in FedDeCAB, allowing clients that are offline from the server to establish collaborative optimization with neighboring clients. As shown in the figure \ref{fig:feddecab-steps}, FedDeCAB first pushes the latest global model to all online clients. Then, the model trained locally by each online client is compared with the global model, and the distribution difference is obtained by calculating the Kullback-Leibler (KL) divergence between the local and global models. A piecewise-defined function is introduced to encourage a fair aggregation process while minimizing the bias and accelerating the convergence of the global model. The function is composed of a quadratic decay function and a linear function to determine the weights for ranking candidate clients for aggregation based on the KL divergence. Clients with higher KL divergence in the initial rounds will have a higher chance of being selected for aggregation. This preference will gradually be ohased out as training proceeds. The ranking weights are also calibrated based on client availability, and this definition is introduced to maintain fair contributions from clients with lower availability. In addition, FedDeCAB enables semi-decentralized communication, allowing offline clients to obtain the model parameters from neighboring clients and calculate the loss on all the neighbor models and also the local model, with randomly sampled local data. If the local data achieve a smaller loss on the neighbor model, the offline client will perform gradient optimization towards the direction of the neighbor model in the parameter space in the next round of local updates. Compared to previous studies, the main contributions of this paper are summarized as follows:

% \begin{itemize}
%     \item We develop a theoretical model to evaluate the data heterogeneity with KL divergence. Then we propose a ranking model considering the effect of availability budgets on FL. Specifically, we use a decaying quadratic function to prioritize learning from statistically heterogeneous clients during initial training rounds. We also consider client participation in finding a balance among clients with varying availability.
%     \item We present FedCAB, an algorithm applying our theoretical model for the probabilistic rankings of the available clients to select in each round of FL model aggregation.
%     \item Empirical studies show the effectiveness and superiority of FedCAB with a limited communication budget and clients that join the learning process in late rounds.
% \end{itemize}

\begin{itemize}
    \item We develop a definition of data availability for federated time-series prediction tasks and propose three scenarios with different data availability distributions. We develop a client connectivity model in these scenarios to define the behavior of clients participating in FL under availability constraints.
    \item We propose FedDeCAB, an algorithm allowing clients offline from the server to communicate with neighboring clients to establish semi-decentralized collaborative optimization. Offline clients calculate the loss of their local data on all neighboring models and make their next round of local optimization towards the model with the smallest loss. To reduce communication overhead, neighboring clients only send the fully connected layers of the model. We adopt the ranking model from the existing work FedCAB to perform probabilistic ranking of online clients for the server to select in each round of FL model aggregation.
    \item Empirical studies are conducted on real-world taxi and vessel trajectory data. The results show the effectiveness and superiority of FedDeCAB under semi-decentralized FL with high offline rate and limited communication budget.
\end{itemize}

\section{Related Work}
% Statistical and system heterogeneity are common yet challenging issues in FL, especially in an edge computing environment with limited availability. We categorize related work into three categories by their approach to the solution.
The common challenging problems in FL time-series prediction mainly contain statistical and system heterogeneity, especially in edge computing environments with limited availability. We categorize the related works into three categories according to their solutions:

% \subsection{FL With Adaptive Aggregation Weights}
% Adaptive weight aggregation methods help prioritize the clients with knowledge prone to be underestimated or ignored, mitigating bias in the FL models and improving the model performance. \citet{reddi2021adaptive} analytically demonstrated the interplay between the number of local training steps and the heterogeneity among clients. AdaFed \cite{tan2022adafed} and F3AST \cite{ribero2022federated} took device participation into account while reducing the learning bias based on client availability.
\subsection{FL With Weighted Client Sampling}
Client sampling is one of the bottlenecks of FL. Reasonable weight sampling strategies help to reduce bias and improve global performance of the model while preserving communication overhead. \citet{chen2022optimal} applies the update norm to measure client importance and approximate the optimal formula for client participation. F3AST\cite{ribero2022federated} considers the dynamic availability of clients and uses an unbiased strategy to improve model performance. The work of \cite{pmlr-v151-jee-cho22a} tends to sample clients with higher local losses to accelerate error convergence. 

\subsection{FL with Constrained Client Availability}
\citet{gu2021fast} proposed a cache-based FL algorithm MIFA, which maintains a record of the client's latest updates on the server side, allowing clients to contribute regardless of their availability. However, the high communication cost imposed by caching the models puts limit on its practical application in real-world edge devices. \citet{bao2023fedcab} proposed FedCAB which quantified client availability and considered the impact of stragglers on the global model. The algorithm of FedCAB dynamically weights clients by considering client participation and local model bias affected by availability. \citet{rodio2023flavail} dynamically adjusts the weights assigned to clients and ignores clients with low availability, thereby balancing the conflicting goals between maximizing convergence speed and minimizing model bias. \citet{zhu2022resilient} proposed a resilient FL scheme that allows edge devices to learn self-distilled neural networks that can be easily pruned to arbitrary sizes, alleviating connection uncertainty by transmitting partial model parameters under faulty network connections. \citet{chen21wireless} explored federated learning schemes under communication constraints of real-world wireless networks. The work of \cite{wu2023fedle} and \cite{arouj22battery} considers the impact of energy consumption on client availability from battery-constrained devices. \citet{yemini22semi} proposed a semi-decentralized method for clients to calculate local consensus of updates from their neighboring clients, improving convergence speed and accuracy.

\subsection{Availability and Heterogeneity with Time-series Data}
% \citet{qinbin2022silos} described the label distribution skew in federated learning as a situation where the distribution of data labels has differences among different clients. Although the algorithm for FedAvg \cite{mcmahan2017communication} is simple and does not require high communication costs, the performance of this algorithm is severely degraded in the situation of label skew. Many solutions have been proposed to tackle the skew problem and improve the model performance. FedProx \cite{tian2022fedprox} introduced a regularization term to the loss function in the local training, where local models distant from the global model in the parameter space will suffer a penalty to improve the stability of the model in a heterogeneous FL scenario. MOON \cite{qinbin2022moon} used a model-contrastive loss term in local loss calculation, simultaneously decreasing the distance between the global model and the local update and increasing the distance between the latest and previous local model. However, all these solutions emphasize the importance of reducing the difference between the global and local models instead of fully considering the impact of limited client availability and extreme label distribution skew.

% While the prior art appreciated the significant impact of client availability, few studies simultaneously consider client availability, communication budgets, and how they interact with client selection and weighted averaging.
In the context of real-world distributed data, time-series prediction based on federated learning can effectively protect the privacy of participants \cite{liu22fedtad}. Time-series data are collected from acquisition systems such as sensors and detectors. Due to the influence of various factors such as objective conditions and the stability of data acquisition equipment, the data often contains outliers such as noise and missing values, which seriously affects the availability of time-series data. In addition, the model accuracy is often affected by high complexity of data, which may also lead to poor generalization performance \cite{liu21forcast} . In the trajectory prediction task, features like the behavior of drivers as well as the complex road conditions naturally bring about the heterogeneity to time-series data \cite{Fang2022HeterogeneousTF}. FedAvg \cite{mcmahan2017communication} is simple and has low communication cost as a FL solution, but its performance will be seriously degraded when data is highly heterogeneous and low available. FedProx \cite{tian2022fedprox} introduces a regularization term to the loss function in local training to penalize local models that deviate far away from the global model in parameter space to improve the stability of the model in heterogeneous FL scenarios. MOON \cite{qinbin2022moon} applies contrastive learning in local loss calculation, reducing the distance between the global model and local updates and increasing the distance between the latest local model and the previous local model. While the prior art recognizes the significant impact of data availability, few studies have simultaneously considered the availability of both data and clients, communication budgets, connectivity, and the mutual influence between clients.

\section{Problem Formulation}

% In this section, we consider the availability constraint of clients and the subsequent implications on the underlying FL process. Then, we present and analyze a novel FL client selection algorithm called FedCAB based on the client availability budgets. In our problem setting, each client has a limited communication budget due to energy constraints. In addition, some clients cannot join the learning process initially but can join later. A typical example is in the edge environments where clients periodically sleep and wake up for energy preservation purposes, missing the initial training rounds. Clients can also be mobile and join the network well after the training begins. 
In this section, we discuss the client availability constraints and their subsequent implications on the underlying FL process. Then, we propose and analyze a novel FL client selection algorithm based on client availability budgets with semi-decentralized collaborative optimization, called FedDeCAB. In our problem setting, there is a limited communication budget with each client due to energy constraints. Additionally, clients have a probability of being offline from the central server in each round. A typical example is that in edge environments, clients miss training rounds because of periodical sleep and wake-up in order to preserve energy; or mobile clients are located in the areas with poor connectivity and lose connection with the server. Offline clients can perform decentralized communication rounds with neighboring clients at a lower frequency, or rejoin the FL network with a certain probability after being offline.

\subsection{Dynamic Client Availability in FL Participation}

% Existing work \cite{gu2021fast} presumed that the likelihood of a client participating in each communication round during the training process remains consistent. 
% This setting is rather restrictive for real-world applications of FL, and we model the participation probability for each client to be dynamic and adjustable. 
% This will match the scenarios where a client may be unavailable due to issues from either the device or the networks in the initial training phases but regain connectivity later, thereby increasing the odds of participating in each round for training and updates. 
% On the other hand, a client may be actively participating in the FL process during the early stages of training but abruptly withdraw in later stages. 
% As such, we formulate the following optimization problem:
Existing work \cite{gu2021fast} assumes that the probability of a client participating in each round of FL communication during training remains consistent.
This setting is quite strict for real-world applications of FL, and we model the participation probability of each client as dynamic and adjustable. This will match the following scenarios: a client may be offline for a period of time due to device or network issues, but will regain connectivity later, increasing the chance of participating in each round of training and updates. In addition, a client may be actively participating in the FL process during the early stages of training, but precipitously drop out in the later stages. Whether offline or not, the client is able to periodically communicate with neighboring clients for collaborative optimization in the semi-decentralized framework of FedDeCAB.
Therefore, we formulate the following optimization problem:

\begin{equation}
    \min_{\mathbf{w} \in \mathbb{R}^d} f(\mathbf{w}) := \frac{1}{N} \sum_{i = 1}^{N} [f_i(\mathbf{w}, D_i, \xi_i, \mathbf{\tau}_i) + g_i(\mathbf{w}, \mathbf{v^*})],
\end{equation}

where $\mathbf{w}$ is an optimizing variable, $f_i$ is the local loss function on the $i$th device, $D_i$ is the local dataset on client $i$, $\xi_i$ describes the local data distribution randomness, and $\tau_i$ is a $T$-dimension vector with binary values describing if the $i$th client is available during a training epoch. Let $\{\nu_i\}$ be the set of model parameters from neighboring clients of $i$th client. $\mathcal{L}(\mathbf{v}, D_i)$ is the loss evaluated for the model parameter $\mathbf{v}$ on data $D_i$, and $g_i$ is the difference between $\mathbf{w}$ and $\mathbf{v^*}$, where $\mathbf{v^*}$ is denoted by:

\begin{equation}
    \mathbf{v^*} = \arg\min_{\mathbf{v} \in {\{\mathbf{\nu}_i}\} \cup \mathbf{w}} \mathcal{L}(\mathbf{v}, D_i),
\end{equation}

\subsection{Unique Data Distribution of Clients}
% In real-world FL datasets, certain edge devices possess data with distinct distributions unique to other devices, a phenomenon called data heterogeneity. Non-i.i.d. datasets commonly exhibit data heterogeneity. Given this context, if devices with unique data distributions have low participation rates during most of the training but experience a brief period of high participation probability, performance might experience a major drop compared to predictable availability. Our proposed idea leverages KL divergence to measure the difference between the last data distribution and the current one locally before sending out the model for caching. In particular, we focus on large data distribution shifts if the clients cannot be sampled for extended epochs. Here, we show the definition of KL divergence below, denoted by $D_{KL}$:
In real-world FL datasets, there is a phenomenon known as data heterogeneity, where some edge devices may have data with distinct distributions unique to other devices. Non-i.i.d. datasets often exhibit data heterogeneity. 
Given this background, if devices with unique data distributions have only a short period of high participation rates during the long training period, performance may drop significantly compared to predictable availability. The idea we propose leverages the KL divergence between model parameters to infer the difference between the previous data distribution and the current data distribution locally before sending the model out. Specifically, we focus on large data distribution shifts if it is not possible to sample the client over an extended period. Here, we show the definition of KL divergence below, denoted by $D_{KL}$:

\begin{equation}
    D_{KL}\left( P || Q \right) = \sum_{i} P_i \ln \frac{P_i}{Q_i},
\end{equation}
where $P$ and $Q$ are the probability distributions to compare; $P_i$ and $Q_i$ are the elements in $P$ and $Q$. In this paper, we take the normalized local update and the previous global model as $P$ and $Q$, respectively, to calculate $D_{KL}$ between those two parameter distributions.

\subsection{Degree of Client Availability}
% Due to device heterogeneity, the availability of clients is affected by network connectivity and user mobility. We assume that all users only have communication constraints, so they consume a limited budget to upload local models. We use $A^k \in [0,1]$ to define the level of participation for Client $k$. The higher $A^k$, the more frequently the client uploads its local model to the server. The definition of $A^k$ at communication round $t$ is shown below:
Client availability is affected by device heterogeneity, mainly characterized by network connectivity and user mobility. We assume that all users only have communication constraints, resulting in a limited budget for uploading their local models. We use $A^k \in [0,1]$ to define the participation level of Client $k$. The higher $A^k$, the more frequently the client uploads its local model to the server. The definition of $A^k$ at communication round $t$ is defined as follows:

\begin{equation}
    A^k = \frac{n^k_t}{t},
\end{equation}
where $n^k_t$ indicates the total number of times by round $t$ the client $k$ uploads its local model to the server.

\subsection{Definition of Trajectory Point Availability}
A trajectory consists of a series of consecutive and ordered coordinate points in the dataset. We define the availability of a trajectory $Tr_i$ with a total number of $n_i$ coordinate points as $\left\{ p^{Tr_i}_j | 0 \leq p_j \leq 1 \right\}_{j \in \left\{ 1, ... , n_i \right\}}$ , where $p^{Tr_i}_j$ represents the probability of the $j-th$ coordinate point of trajectory $Tr_i$ joining the available training dataset of the  corresponding FL client. In order to simulate the real-life behavior of vehicles obtaining coordinate information through positioning systems in real-time prediction tasks, we slice the trajectory data held by each FL client, where the size of each slice is equal to the batch size multiplied by the input trajectory length for the LSTM model. At the beginning, the available training dataset for each client is a very small subset of its original data possessed. As the number of federated communication rounds increases, each client will gradually obtain new coordinate points in the available training data set. For each coordinate point in the trajectory $Tr_i$, the probability that the client adds the $j-th$ coordinate point to the available training dataset is $p^{Tr_i}_j$. Any coordinate points that fail to  join the available training dataset will be permanently unavailable throughout the FL training. The client persists all coordinate points in the available training dataset and can only use these available points for local updates. In particular, we summarize the following 3 scenarios that introduce trajectory availability:

\subsubsection{Random Trajectory Availability}
Assume that when the client device is located in an area with moderate traffic flow, such as offshore waters or urban areas, the positioning signal can be sent and received normally except for occasional random fluctuations. In this case, we can regard each data point in its trajectory as having availability, which is independently subject to random distribution. We generate the availability of each trajectory point through Dirichlet distribution. The trajectory availability based on Dirichlet distribution is denoted as:

$\left\{ p^{Tr_i}_j | p_j \sim Dirichlet(\alpha) \right\}_{j \in \left\{ 1, ... , n_i \right\}}$

\subsubsection{Region-based Trajectory Availability}
Sometimes vessels sail across multiple sea areas from inshore to offshore, and vehicles travel between urban areas and remote suburbs. In this case, the ability of client devices to navigate through base stations will be affected by traffic congestion or regionally weak signal coverage, thus showing varying degrees of decreased availability. For simplicity, we manually divide a series of weak signal areas in the entire trajectory dataset, and the availability of coordinate points in these areas will be relatively lower. The region-based trajectory availability is denoted as:

$\left\{ p^{Tr_i}_j = p_{low} | \forall Tr^j_i \in \left\{ Weak Signal Area \right\} \right\}_{j \in \left\{ 1, ... , n_i \right\}} \cup \left\{ p^{Tr_i}_k = p_{high} | \forall Tr^k_i \notin \left\{ Weak Signal Area \right\} \right\}_{k \in \left\{ 1, ... , n_i \right\}} $

\subsubsection{Data Size Based Trajectory Availability}
Freight or shipping company vessels travel frequently and usually have more available tracks than privately owned smaller ships. To ensure shipping safety and improve operational efficiency, company vessels usually have better transceiver equipment for positioning signal . The same applies to vehicle operating companies and private cars. In this scenario, trajectory data is only split based on the vehicle ID in the dataset, with each client holding all data from one vehicle. Since the data of different device IDs account for different proportions in the entire dataset, this will lead to a strong heterogeneity in the client data size. Clients with more coordinate points are regarded as company devices and are assigned higher trajectory availability. Clients with fewer coordinate points are considered as private or small devices, whose availability will be set lower. By adjusting the size threshold between company and private devices, we make a small number of clients have high data availability, while a large number of clients have lower data availability, thereby creating a more heterogeneous availability setting. For this scenario, assuming that the trajectory $Tr_i$ belongs to client $C$, the trajectory availability based on data size is expressed as:

$\left\{ p^{Tr_i}_j | p_j = AssignAvailability(C) \right\}_{j \in \left\{ 1, ... , n_i \right\}} $

\section{The FedDeCAB Algorithm}

In this section, we formally introduce the FedDeCAB algorithm for weighted client selection and semi-decentralized collaborative optimization, shown in Algorithm \ref{alg:FedCAB} and Algorithm \ref{alg:FedDeCAB}, where the central server will assign a weight to each available client based on 1) how actively they participate in the communication, and 2) a decaying KL divergence loss between their local update gradients and the global model. Meanwhile, offline clients will perform decentralized collaborative optimization with neighboring clients.

\subsection{Availability-budget-based Client Selection Strategy}
In each round, the server applies the above two strategies and uses a probabilistic method to rank all available clients, then sampling clients with higher ranking weights for update.

The server evaluates client participation by calculating the frequency of each client's participation in model updates. Clients with lower participation frequencies are given higher weights, which promotes the global model to learn from rarely extracted knowledge. We adopt a series of quadratic weight functions for $D_{KL}$. Initially, clients with higher $D_{KL}$ are given higher weights to avoid large biases in the global model. As the number of global communication rounds increases, the weight function will gradually decay to a linear function. After this stage, clients with lower $D_{KL}$ will be given higher weights to accelerate the convergence of the global model.

Let $N$ denote the total number of clients in the network. We use $i$ to refer to the $i$th client. Let \emph{T} be the total number of global rounds. Define \emph{K} as the number of clients sampled per round. Suppose \emph{E} is the number of client local training epochs. $\{\eta\}^\emph{T}_{\emph{t} = 1}$ is the learning rate schedule. $\alpha $ is the compensator for clients with high KL-divergence loss. $\Delta\alpha$ is the decay value of $\alpha$ per round. $\gamma$ is the compensator for straggling clients.

Define $m_t$ as the total number of clients participating in the ranking at round $t$. Assume $P^k \in [1, m_t]$ is the overall ranking of client $k$ among all participating clients when computing $D_{KL}$ between its local update and the global model at round $t$. We further define $b_0$, $b_1$, and $b_2$ to be the three parameters of a concave quadratic curve calculated by solving the three points $(0, \alpha)$, $(m_t, 1)$ and $(2 m_t, \alpha)$ on the curve.
Define a compensation parameter $\alpha$ and its per-round weight decay of $\Delta\alpha$. 
In the early stage of training, the weight distribution $G^k_t$ for Client $k$ at Round $t$ can be denoted by:

\begin{equation}
    G^k_t (\alpha > 1)= b_0{{P}^k}^2 + b_1{P}^k + b_2,
    \label{eq:gtk}
\end{equation}
In Equation (\ref{eq:gtk}), $b_0$, $b_1$ and $b_2$ are solved by:

\begin{align}
& b_0 = \frac{\alpha -1}{{m_t}^2}, b_1 = \frac{-2m_t(\alpha -1)}{{m_t}^2}, b_2 = \alpha.
\label{eq:b012}
\end{align}
In Equation (\ref{eq:b012}), $\alpha$ will decay with its weight each round as $\alpha_{t+1} \gets \alpha_{t} - \Delta\alpha$. When $\alpha$ decays to value such that $\alpha \leq 1$, the quadratic curve weight allocation will no longer be applicable. The server will instead use a linearly increasing weight assignment:

\begin{equation}
    G^k_t (\alpha \leq 1) = \frac{{P}^k}{m_t}.
\end{equation}

\renewcommand{\algorithmicrequire}{\textbf{Input:}}
\renewcommand{\algorithmicensure}{\textbf{Output:}}

\begin{algorithm}[htb]
    \caption{FedDeCAB: Centralized FL Optimization}
    \begin{algorithmic}[1]
    \REQUIRE $\emph{T}, \emph{K}, \emph{N}, \emph{E}, \{\eta\}^\emph{T}_{\emph{t} = 1}, \alpha, \Delta\alpha, \gamma $
    \ENSURE Global model $\textbf{$\bar{W}$}_\emph{T} $
    \STATE Initialize $\textbf{$\bar{W}$}_\emph{0} $
    \STATE Initialize $\beta^k, \Delta\beta^k$ for each client $k \in \emph{N}$
    \FOR{$\emph{t} = 1\rightarrow\emph{T}$}
        \STATE $C_t \gets $ set of available clients at round $t$
        \STATE $R_t \gets $ set of recovered clients at round $t$
        \STATE $m_t \gets $ number of available clients at round $t$
        \FOR {each client $k \in C_t$ in parallel}
            \IF{$k \in R_t$}
                \STATE ${w}^k_{t+1} \gets $ LocalUpdate$ \left( k, \textbf{${w}^k$}_{t}, E, {\eta}_t, {\nu}^k_t \right)$
            \ELSE
                \STATE ${w}^k_{t+1} \gets $ LocalUpdate$ \left( k, \textbf{$\bar{W}$}_{t}, E, {\eta}_t \right)$
            \ENDIF
            \STATE ${L}^k \gets D_{KL} \left( \textbf{$\bar{W}$}_{t}, \textbf{w}^k_{t+1} \right)$
            \STATE ${n}^k \gets $ total number of updates
            \STATE ${A}^k \gets \frac{n^k_t}{t}$
        \ENDFOR
        \FOR {each client $k \in C_t$ , the server}
            \STATE ${P}_L^k , {P}_A^k \gets $ position of ${L}^k$ in $\left\{ L \right\}$ and ${A}^k$ in $\left\{ A \right\}$ sorted from high to low
            \IF{$\alpha > 1$}
                \STATE $b_0, b_1, b_2 \gets$ Solve Eq. (\ref{eq:b012}) with $\left( \alpha, m_t \right)$
                \STATE $R^k_t \gets \frac{\left(b_0{{P}_L^k}^2 + b_1{P}_L^k + b_2\right){P}_A^k}{m_t}\beta^k $ 
            \ELSE
                \STATE $R^k_t \gets \frac{{P}_L^k{P}_A^k}{{m_t}^2}\beta^k $
            \ENDIF
            \IF{${n}^k < \frac{\begin{matrix} \sum_{i \in \left\{ n \right\}_t} {n}^i \end{matrix}  }{m_t}   $}
            \STATE $R^k_t \gets \gamma R^k_t $ 
            \ENDIF
            \STATE $\beta^k \gets $ Max$ \left(\beta^k - \Delta\beta^k, 1 \right)$
            
        \ENDFOR
        \FOR{$\emph{i} = 1\rightarrow$ min $\left( K, m_t \right)$}
            \STATE $r^i_t \gets$ the $i_{th}$ client in $\left\{ R_t \right\}$ sorted from largest to smallest value
            \STATE $r^i_t$ sends ${w}^i_{t+1}$ to server, and updates its activeness
        \ENDFOR
        \STATE $\textbf{$\bar{W}$}_{t+1} \gets $ ServerUpdate$ \left( \textbf{$\bar{w}$}^{\left\{ r \right\}}_{t+1}, \left\{ r \right\} \right)$
        \STATE $\alpha \gets \alpha - \Delta\alpha$
        \IF{$t$ in decentralized communication rounds}
            \STATE Perform Decentralized Optimization Round
        \ENDIF
    \ENDFOR
    \end{algorithmic}
    \label{alg:FedCAB}
\end{algorithm}

\begin{algorithm}[htb]
    \caption{FedDeCAB: Decentralized Optimization Rounds}
    \begin{algorithmic}[1]
    \REQUIRE $\emph{E}, \emph{t}, \{\eta\}^\emph{T}_{\emph{t} = 1}, \Omega_t $
    \ENSURE $\{w^u_{t+1}\} $
    \STATE $\Omega_t \gets $ set of offline clients at round $t$
    \FOR {each client $u \in \Omega_t$ in parallel}
        \STATE $\nu^*_u \gets u$ 's best collaborative model from last round
        \STATE ${w}^u_{t+1} \gets $ LocalUpdate$ \left( u, \textbf{${w}^u$}_{t}, E, {\eta}_t, {\nu}^*_u \right)$
        \STATE $\{\nu_i\}^\emph{$\chi$}_{\emph{i} = 1} \gets $ set of model parameters from $\chi$ neighboring clients
        \STATE $\nu^*_u \gets $ $u$ finds a $\nu \in {\{\mathbf{\nu}_i}\} \cup {w}^u_{t+1}$ which minimizes $\mathcal{L}(\mathbf{\nu}, D_u)$
    \ENDFOR
    
    \end{algorithmic}
    \label{alg:FedDeCAB}
\end{algorithm}

\subsection{The FedDeCAB Algorithm: Weighted Client Selection}
Now we discuss the details of the pseudocode in algorithm \ref{alg:FedCAB}. The client ranking strategy of FedDeCAB is designed to compensate the participation opportunities of clients with lower participation using a weight criterion that is universally applicable to all clients, while making the global model move towards a better solution in the early stage of training. In order to achieve this, the server first sets up the KL-divergence compensation parameter $\alpha$, its per-round attenuation value $\Delta\alpha$, and the straggler compensation $\gamma$. At the same time, all clients initialize $\beta$ and its attenuation $\Delta\beta$ (Line 2). $\beta$ gradually decays as the number of updates increases, and is used as a compensator for late-joining clients. For each client, $\beta$ and $\Delta\beta$ can be initialized separately by either the server or the client device, because each client's $\beta$ only affects the ranking weight of the client itself. For fairness, we only discuss the case where the initial value $\beta$  among clients is exactly the same in this paper, and so is $\Delta\beta$. Specifically, $\beta$ is multiplied as an additional coefficient on the client's ranking weight. The initial value of $\beta$ is greater than 1, so the straggler clients tend to get higher ranking weight modifications. $\beta$ is not affected by the KL divergence calculation or client participation, but only by its decay weight $\Delta\beta$. When $\beta$ is less than 1, it will no longer be in effect.

At each round $t$, the server obtains a subset $C_t$ from the $m_t$ available clients. For clients that were previously offline but are online in this round, we name them recovered clients, denoted by $R_t$. The server sends the latest global model to all clients in $C_t$ that are not in $R_t$. Then, each client $k$ in $C_t$ performs a local update on the local model ${w}^k_{t+1}$. It is worth noting that the recovered client will use the model cached in the previous decentralized communication round for collaborative optimization (Line 9). The clients then calculate the KL divergence between the received global model and their updated local model. At this stage, these available clients only send the KL divergence value, its availability history, and the number of times it has updated with the server (Lines 7-15).

After collecting the model updates from these clients, the server generates two priority queues for the clients ranked by client participation and the KL divergence calculation uploaded by each client (Line 18). If the compensator $\alpha$ is in effect (Line 19), the server substitutes three points ($0, \alpha$), ($m_t, 1$), and ($2m_t, \alpha$) to solve the unique quadratic function and obtain its parameter set ($b_0, b_1, b_2$) (Line 20). The quadratic function gives higher weights to clients with high KL divergence (Line 21). As $\alpha$ gradually decays to 1, this quadratic function approaches $y=1$. When $\alpha \leq 1$, compensation for high KL divergence will no longer take effect. Thereafter, clients with lower KL divergence will receive more weight in the ranking to speed up model convergence (Line 23). $\beta$ is another compensation coefficient for newly joined clients. Each client has a higher weight modifier in the first few updates, aiming to extract the contribution of late-joining clients. This compensation will decay as the client updates and gradually decrease to less than 1, at which point it will no longer be in effect. $\gamma$ is a server parameter used to boost the weights of stragglers and late-joining clients when its value is larger than 1 (Lines 25-26).
The modified ranking weight $R^k_t$ of each available client $k$ is used for the final ranking, where the top-ranked subset of clients is sampled based on the number of clients $K$ per round. These sampled clients then upload their local updates to the server. The server then aggregates the local solutions to obtain a global model for the next round (Lines 30-34).

The frequency of decentralized rounds where offline clients communicate with their neighboring clients will be lower, usually performed once after several centralized FL communication rounds. This is designed to preserve client communication budgets (Lines 36-37).

\subsection{The FedDeCAB Algorithm: Decentralized Collaboration}
The pseudocode of FedDeCAB decentralized communication and collaboration is described in algorithm \ref{alg:FedDeCAB}. The decentralized rounds are designed for clients that temporarily lose connection with central server, namely offline clients. Despite the communication limitations with the central server, offline clients still have adequate capability to perform local computation. More importantly, these clients have chance to perform lightweight communications with other clients that are geographically close, providing opportunity for decentralized collaborative optimization. 

In the $t$th decentralized communication round, each client $u$ from the offline clients set $\Omega_t$ performs local updates with the best collaborative model $\nu^*_u$ cached in the previous offline round to obtain the new model parameters ${w}^u_{t+1}$ (Lines 3-4). Then, client $u$ obtains local model parameters from $\chi$ neighboring clients within the communication range as candidate collaborative models (Line 5). Then, client $u$ calculates the output embedding from the LSTM layer, applying it respectively to the FC layers of candidate collaborative models and its own. The loss values are calculated after the embedding pass through all the candidate models. Eventually, the model parameters that obtain the minimum loss will be selected as the new best collaborative model of client $u$ (Line 6). When the client are using its best collaborative model $\nu^*_u$ to perform local updates, in the optimization function there will be an extra bias term based on the empirical loss to penalize the local model parameters for deviating from the collaborative model parameters. Specifically, we consider the parameters of the FC layers as a distribution, and calculate the difference between the two distributions as the bias term. The loss function is calculated by: 

\begin{equation}
    f(\mathbf{w}) := \mathcal{L}(\bar{y}, y) + D_{KL}({w}, \nu^*)
\end{equation}

\begin{figure*}[!htb]
    \begin{minipage}{0.35\textwidth}
        \centering
        \includegraphics[scale=0.025]{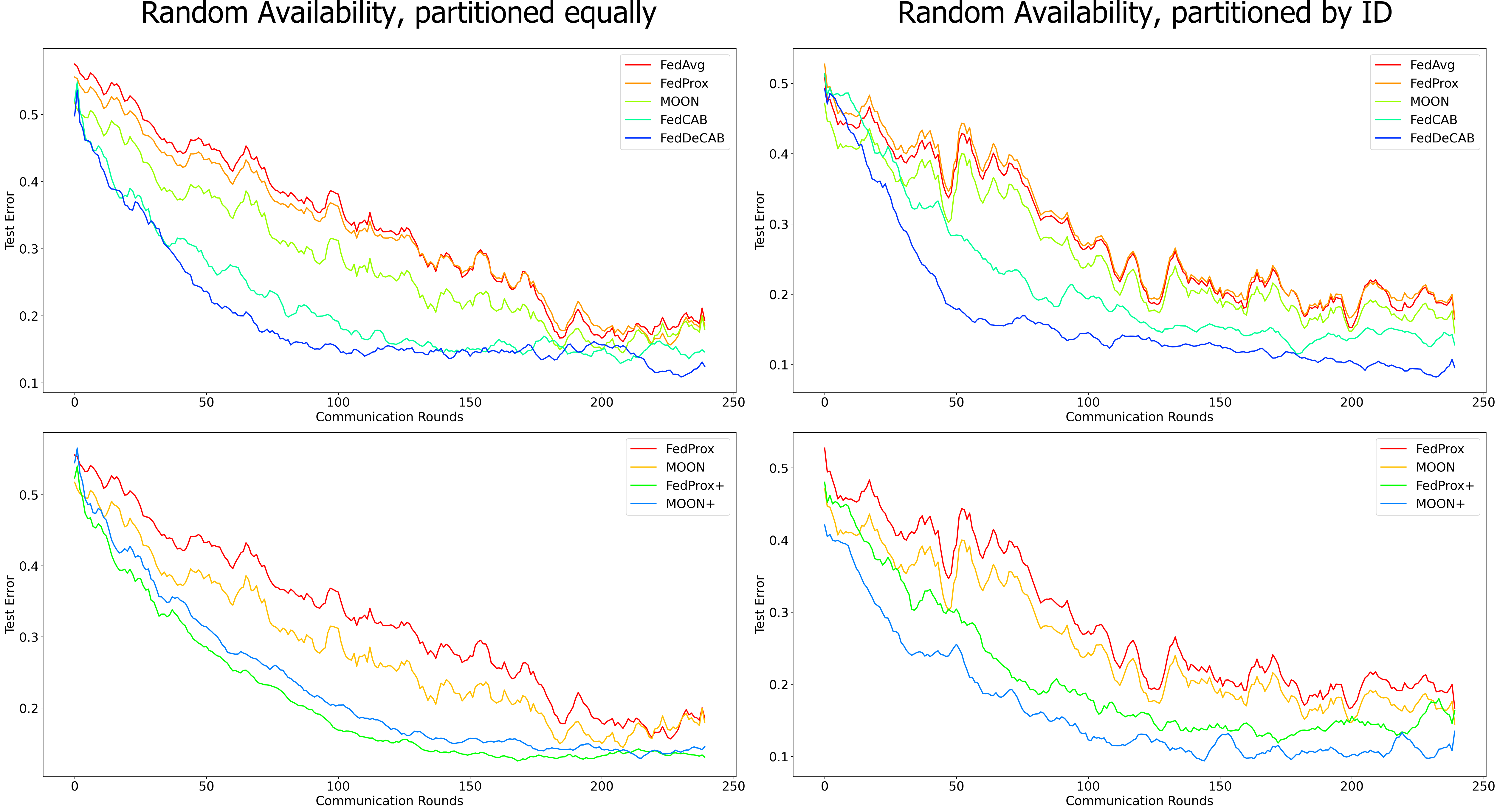}
        \caption{Random availability for Vessel trajectory dataset.}
        \label{fig:collation-vessel-random}
    \end{minipage}\hfill
    \begin{minipage}{0.35\textwidth}
        \centering
        \includegraphics[scale=0.025]{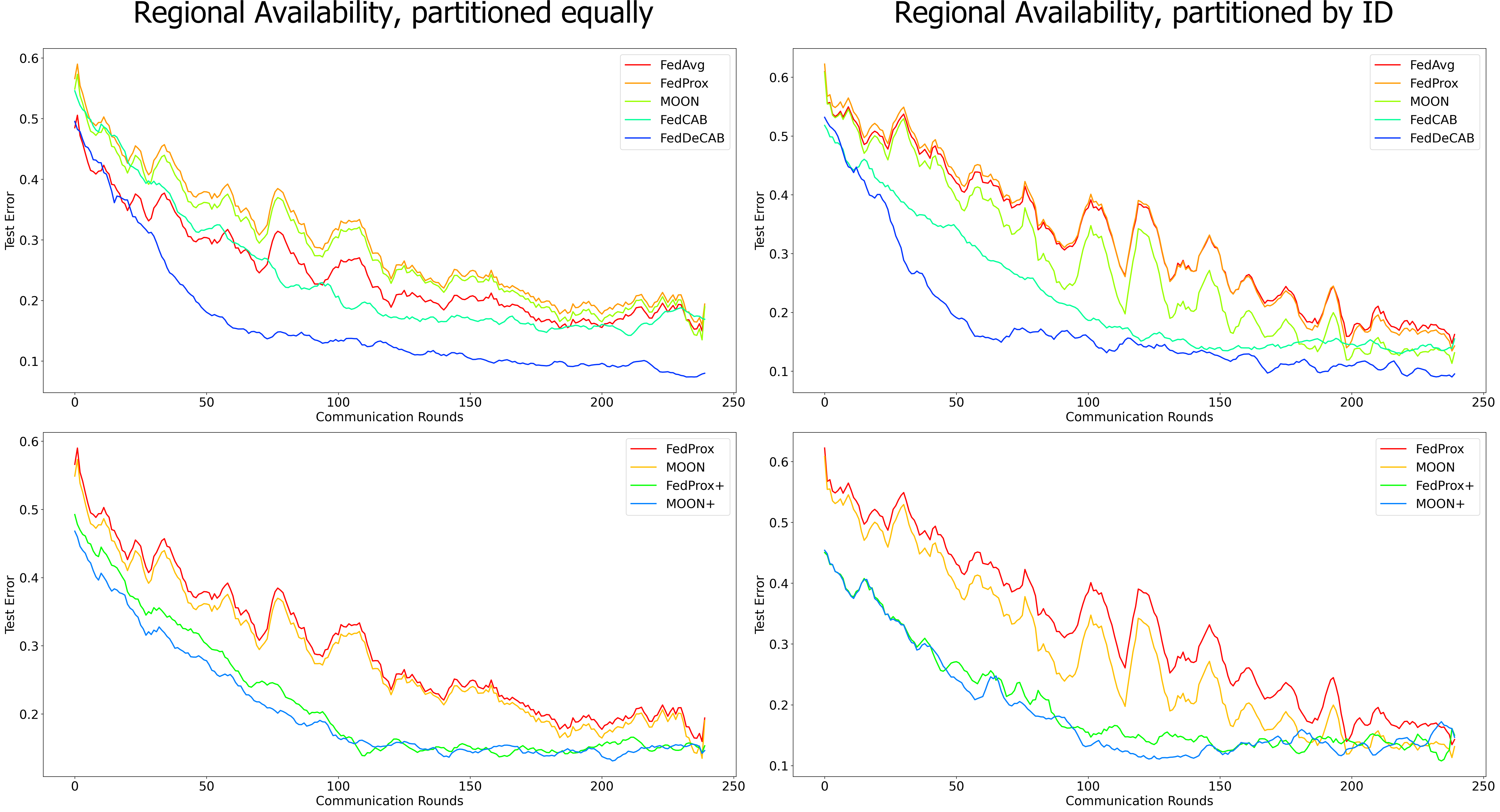}
        \caption{Regional availability for Vessel trajectory dataset.}
        \label{fig:collation-vessel-regional}
   \end{minipage}\hfill
   \begin{minipage}{0.22\textwidth}
        \centering
        \includegraphics[scale=0.025]{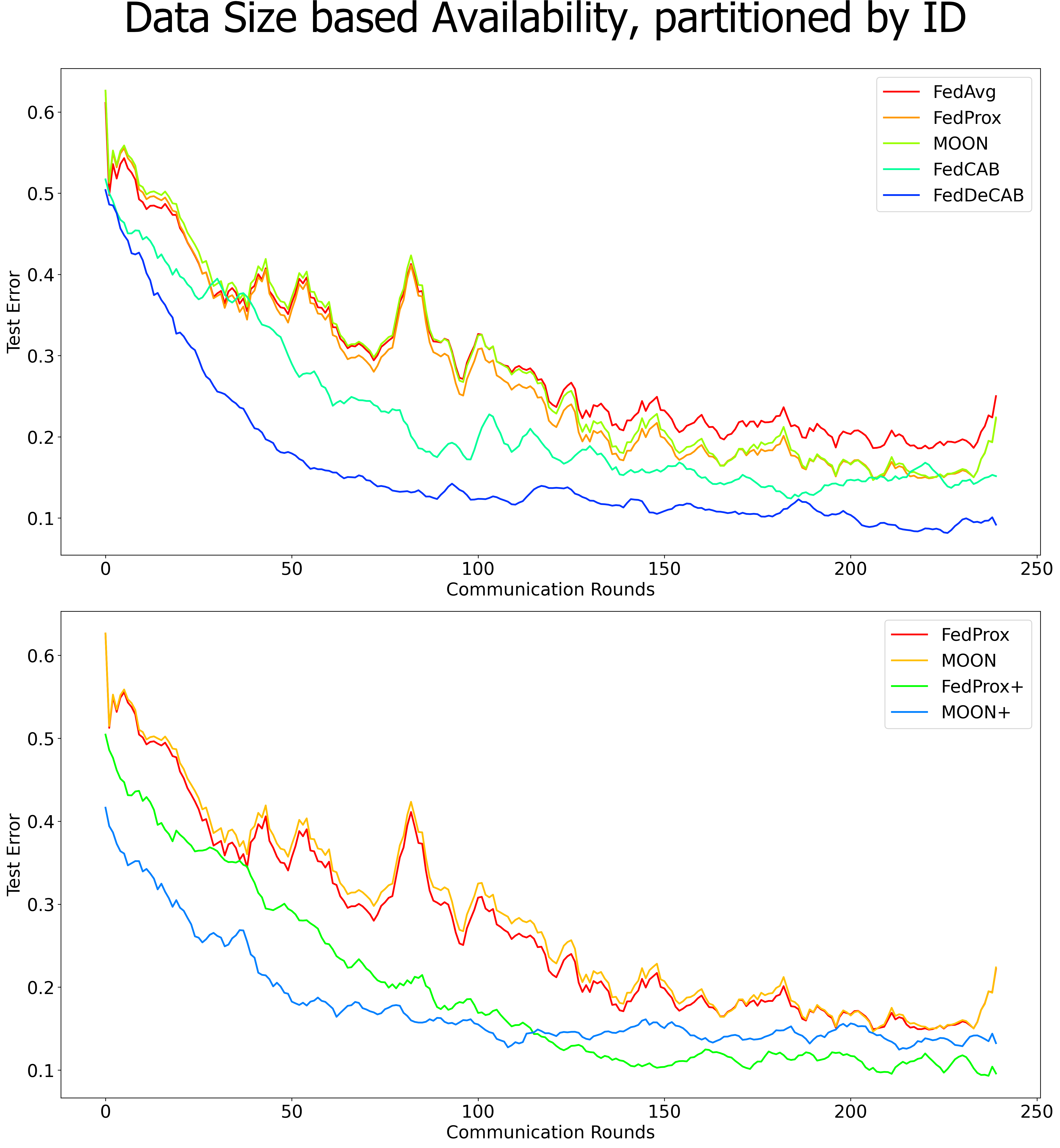}
        \caption{Data size based availability for Vessel trajectory dataset.}
        \label{fig:collation-vessel-datasize}
    \end{minipage}\hfill
\end{figure*}

\begin{figure*}[!htb]
    \begin{minipage}{0.35\textwidth}
        \centering
        \includegraphics[scale=0.025]{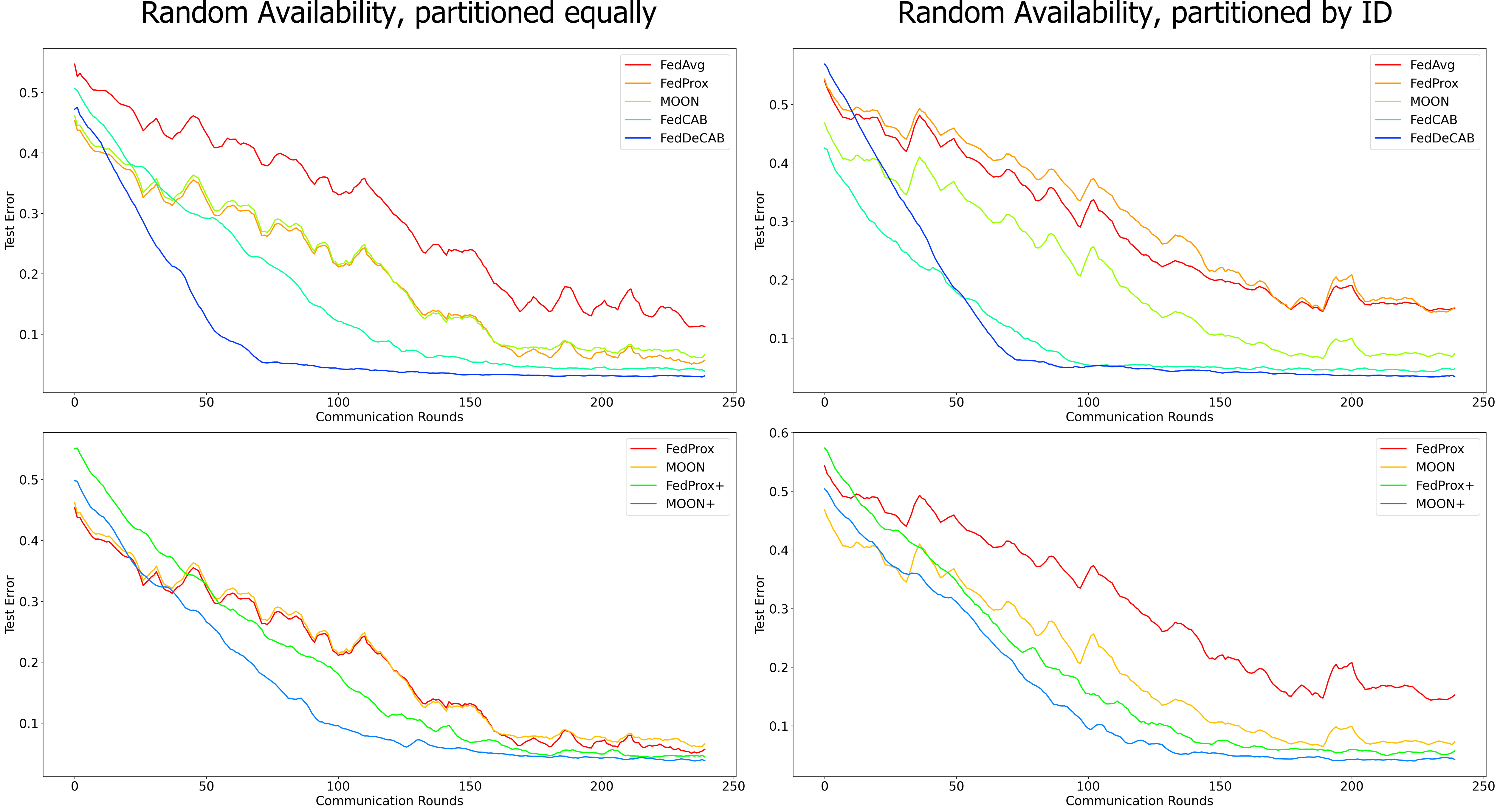}
        \caption{Random availability for T-Drive dataset.}
        \label{fig:collation-tdrive-random}
    \end{minipage}\hfill
    \begin{minipage}{0.35\textwidth}
        \centering
        \includegraphics[scale=0.025]{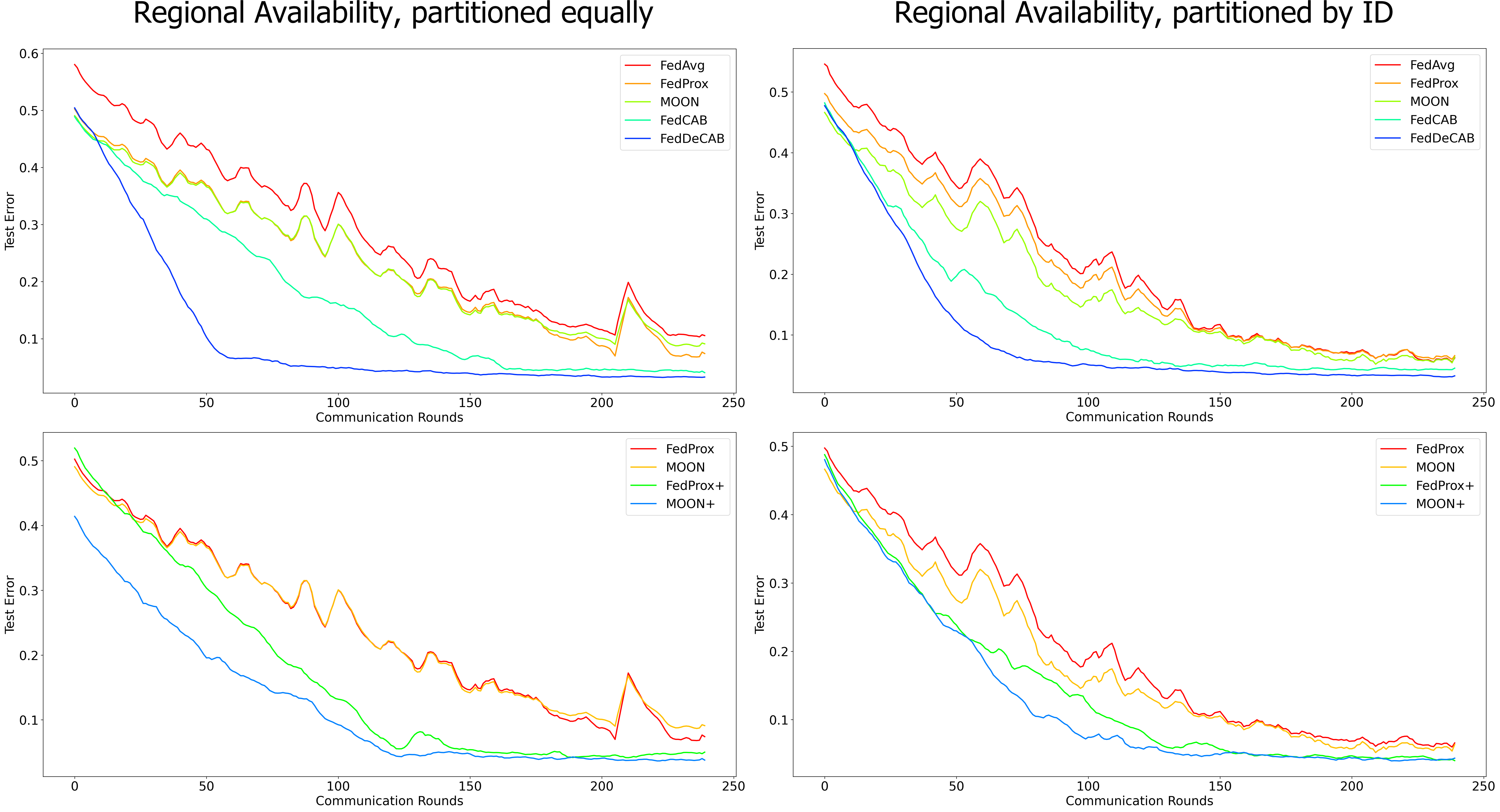}
        \caption{Regional availability for T-Drive dataset.}
        \label{fig:collation-tdrive-regional}
   \end{minipage}\hfill
   \begin{minipage}{0.22\textwidth}
        \centering
        \includegraphics[scale=0.025]{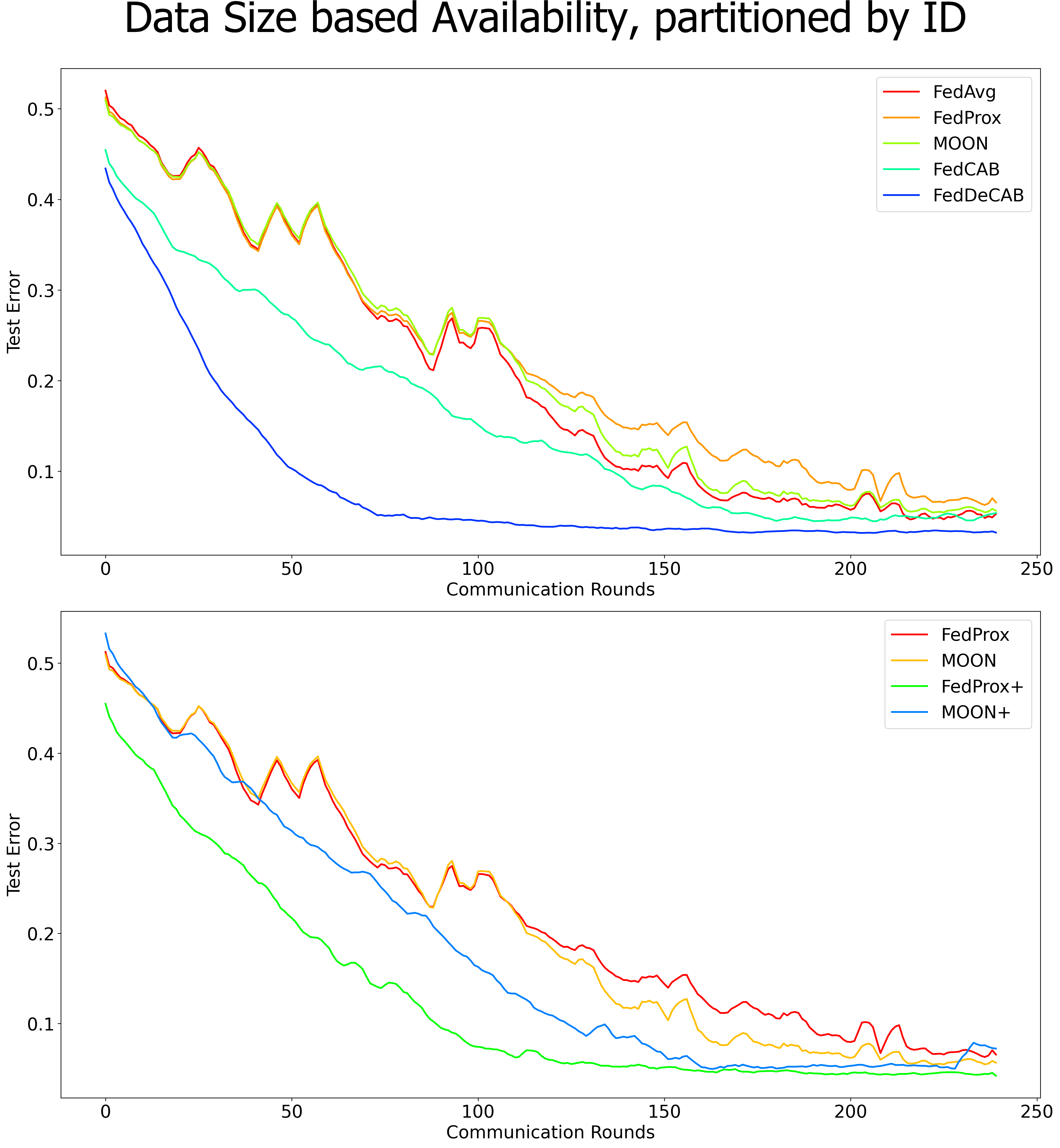}
        \caption{Data size based availability for T-Drive dataset.}
        \label{fig:collation-tdrive-datasize}
    \end{minipage}\hfill
\end{figure*}

\section{Numerical Results}
In this section, we evaluate our algorithms and investigate how the communication budget constraint and straggling devices influence the performance of federated learning. 

\subsection{Experimental Settings}
We compare our algorithm with the following three method: FedAvg \cite{mcmahan2017communication}, FedProx \cite{tian2022fedprox}, MOON \cite{qinbin2022moon} and FedCAB \cite{bao2023fedcab}. In addition, we also extend the work of FedCAB and combine its framework of client ranking selection respectively with FedProx and MOON, namely FedProx+ and MOON+. We use a LSTM model with 128 units, 6 time-steps and a batch size of 16. We tested our algorithm and all baselines on a vessel trajectory dataset of China sea \cite{vesseldata} and T-Drive taxi trajectory dataset \cite{zheng2011t-drive}. Data is partitioned either equally or by vehicle ID. When partitioned equally, each client in vessel trajectory dataset gets 2500 data points, and 18900 each for clients in T-Drive. When partitioned by vehicle ID, client gets 1 vessel of data from vessel trajectory dataset or 4 vehicles of data from T-Drive. Each trajectory in the dataset is unique and does not overlap other trajectories, forming a natural heterogeneous non-i.i.d. data distribution. The number of trajectories differs from vehicle to vehicle, resulting in an extreme data distribution skew. 

To ensure the fairness of the experiment, all algorithms will use the same sampling ratio per round as $m_t = 10\%$. Clients are sampled uniformly and randomly in each baseline algorithm, and we use $E$ = 1 local epoch throughout the experiments. We set the learning rate as $\eta_0 = 0.001$. To ensure the fair contribution among all clients, the initial communication budget is set to be the same for every client, which means each client's local model will have the same opportunity participating in the server's global aggregation throughout the experiment. We use two different data partitioning methods: The first method is to assign all trajectory points of a vehicle to the same client according to the ID; The second method is to allocate to each client an equal number of trajectory points that are as consecutive as possible. We used root mean squared error (RMSE) as the experimental evaluation metric, which is denoted by:

\begin{equation}
    RMSE(y, \hat{y}) = \sqrt{\frac{\sum_{i=1}^{n} (y_i - \hat{y}_i)^2}{n}}
\end{equation}
where $n$ is the total number of data points; $y$ and $\hat{y}$ indicates the ground truth and the predicted value, respectively.

\subsection{Impact of Offline Clients}
In real-time federated learning scenarios, some clients disconnect from the server and rejoin after several rounds. This phenomenon occurs frequently when clients are affected by heterogeneity in multiple aspects. Due to the data heterogeneity between different clients, and the uniqueness of data in the trajectory dataset will naturally amplify this effect, the local optimization goals of these late-joining clients may be inconsistent with the global optimization goals, which will have a negative impact on federated optimization. Offline clients cannot communicate with the server, which means they are not available in FL aggregation. However, offline clients can still try to train local models through collaborative optimization with geographically neighboring clients. We observe their impact on model accuracy through different data availability scenarios and data partitioning.

\begin{figure}[tb]
\centering
\includegraphics[scale=0.11]{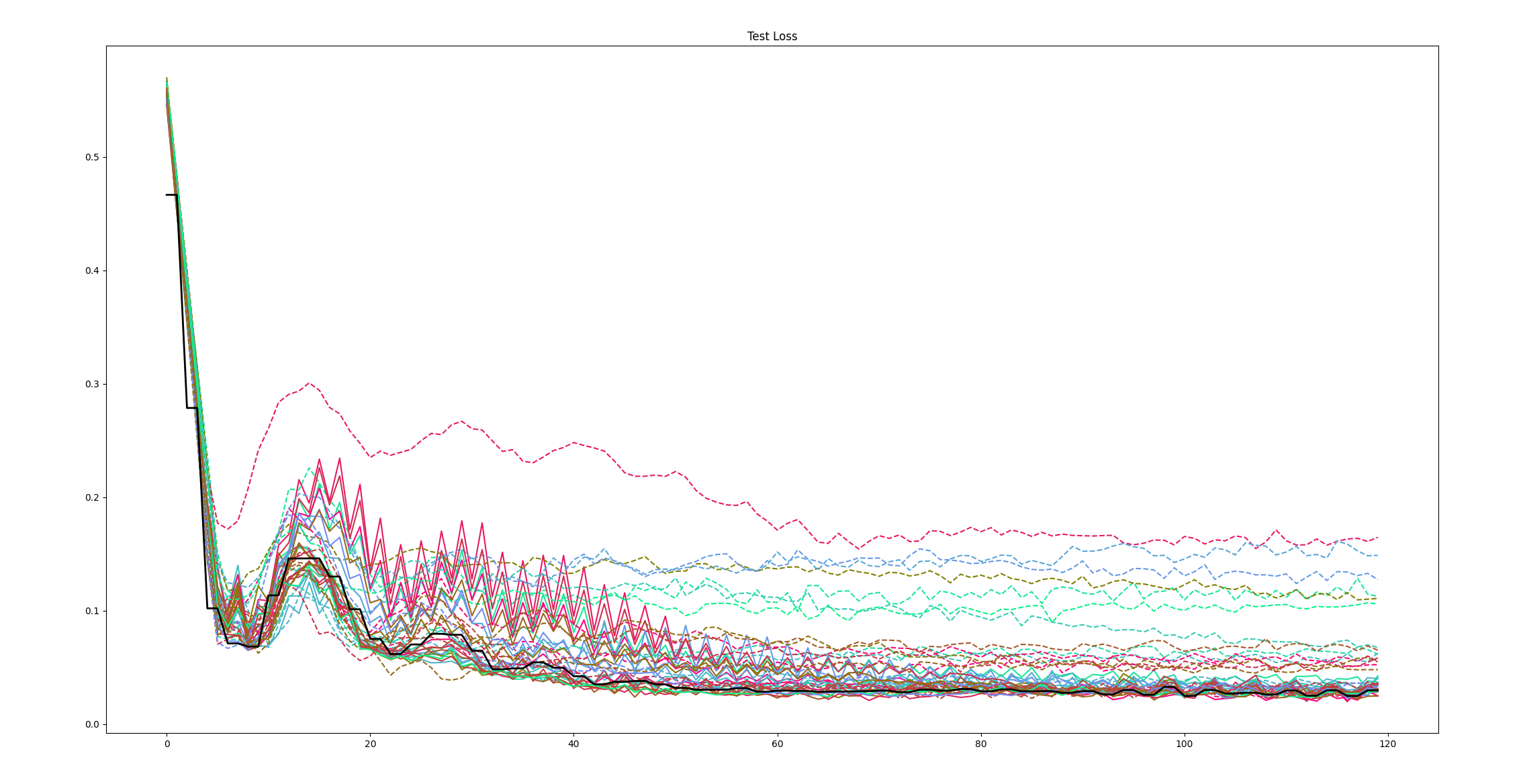}
\caption{Federated aggregation versus local-only training: The colored solid lines and dashed lines respectively indicate the model test loss of FL clients and local-only clients in each training iteration. The black dashed lines indicate the test loss of the FL aggregated model.}
\label{fig:local-only-training}
\end{figure}

\subsection{Federated aggregation versus local-only training}
In this section, we experimentally verify the performance improvement of federated learning compared to local training without aggregation in the time-series trajectory prediction task. We use FedAvg as the federated learning method, a standard baseline for FL, which samples available clients with a normalized probability and uses SGD as the server optimizer. For local-only clients, they will train the model only based on their local data, and will not share any information with other clients. We set 10 clients respectively for FL training and local training. The training round is 60, and there are 2 iterations for local update in each round, which means both FL clients and local clients will have 120 times of model update. The availability is set to 1.0 for all trajectory points, which means clients are allowed to use all of their training data. The result is shown in Fig. \ref{fig:local-only-training}. We observe that FL clients have a better performance than clients training only with their own local data. We believe that federated learning enables clients to share knowledge with each other, making up for the knowledge deficiency of local data.
% We use the same data distribution for each experiment, and the communication budget for each client stays the same. Out of the 100 clients, we choose 20 as late-join clients, with unique digit classes $0$ and $1$ that the other clients do not have. Late-join clients are unavailable in the early stage of training, which means they can neither communicate with the server nor train the local model by themselves. These 20 clients become available after 50 rounds. We watch their impact on model accuracy by varying the number of rounds they wait before joining the training, and the results are shown in Fig. \ref{fig:exp-late-join-fedavg} and Fig. \ref{fig:exp-late-join-fedcab}. We observe that for FedAvg, the later the late-join client joins the training, the worse the final performance of the global model is. We believe that the client who joins the training earlier dilutes the contribution and influence of the late-join clients, which prevents the global model from effectively developing toward the optimal solution. On the other hand, our method is overall better than FedAvg and can even improve more when the severity of client straggling is moderate.

\subsection{Exploration on different trajectory availability settings}
We further investigate the impact of trajectory availability on different FL algorithms. We set the total number of clients $N$ = 40, with the client availability budgets set to 20. The training round $T$ is 240. The probability of an online client going offline each round is 0.2, and the probability of an offline client restoring FL centralized communication each round is 0.1. For collaborative optimization among offline clients, we set the frequency of decentralized rounds as $50\%$ of the regular FL centralized rounds. From the perspective of preserving communication overhead, we let neighboring clients only transmit parameters of fully connected (FC) layers, which hold about only 0.8$\%$ (645 parameters) of the total LSTM model (81408 parameters) used in the paper.

In this experiment, we evaluate the ranking-based client selection strategy and semi-decentralized collaborative optimization framework  of FedDeCAB on two datasets. We test the performance of FedDeCAB and the baselines in three aforementioned trajectory data availability scenarios: random availability, regional availability, and availability based on data size. In addition, we also test the performance of FedProx+ and MOON+ relatively to the corresponding original algorithms in these scenarios.

The experimental results on the vessel trajectory dataset are shown in Figures \ref{fig:collation-vessel-random}, \ref{fig:collation-vessel-regional} and \ref{fig:collation-vessel-datasize}. Results on T-Drive dataset are shown in Figures \ref{fig:collation-tdrive-random}, \ref{fig:collation-tdrive-regional} and \ref{fig:collation-tdrive-datasize} We observe that FedAvg, FedProx, and MOON based on random sampling client selection strategies show large volatility and poor performance under the circumstances of constrained data availability and unstable client connectivity. FedCAB lacks the ability to extract knowledge from offline clients and usually converges slowly. The weighted ranking-based client selection method combined with the collaborative optimization framework FedDeCAB can effectively extract the contribution of offline clients to the global model and accelerate model convergence, thereby further improving the model performance. In addition, MOON+ and FedProx+, which combine the strategy of client ranking selection from FedCAB, also show better convergence and less model error than the original algorithms using random client sampling.

\section{Conclusions}
% This paper explored the impact of availability budgets on federated learning in IoT clients, including participation timing and energy-limited communication budgets. Through theoretical analysis, we applied a decaying quadratic function to prioritize early training with diverse clients, considering availability to balance client participation. We introduced FedCAB, an algorithm to rank and select clients in each FL aggregation round probabilistically. Numerical results confirmed FedCAB's effectiveness and superiority in scenarios with label distribution skew, limited communication budgets, and late-joining clients. Future work includes using a cache-based mechanism to assist a communication-budget-constrained FL and improve its performance.
This paper explored the impact of availability budget and client offline on the federated learning time-series prediction task, including the data availability, energy-constrained communication budgets and unstable client connectivity. Through theoretical analysis, we applied a decaying quadratic function to prioritize distribution heterogeneity of different clients and consider availability to balance client participation. In addition, offline clients performed decentralized collaborative optimization with neighboring clients using lightweight communication. We introduced FedDeCAB, a framework that includes a probabilistic client ranking and selection algorithm for centralized FL aggregation and cooperative optimization for semi-decentralized communication. Numerical results confirm the effectiveness and superiority of FedDeCAB in various scenarios with high data heterogeneity, limited communication budget, and frequent client disconnections. Future work includes the development of more robust decentralized communication mechanisms to improve the performance of FL clients with communication budgets constraints and connectivity issues.

\bibliographystyle{IEEEtranN}
\bibliography{refs}

% Generated by IEEEtranN.bst, version: 1.14 (2015/08/26)
\begin{thebibliography}{30}
\providecommand{\natexlab}[1]{#1}
\providecommand{\url}[1]{#1}
\csname url@samestyle\endcsname
\providecommand{\newblock}{\relax}
\providecommand{\bibinfo}[2]{#2}
\providecommand{\BIBentrySTDinterwordspacing}{\spaceskip=0pt\relax}
\providecommand{\BIBentryALTinterwordstretchfactor}{4}
\providecommand{\BIBentryALTinterwordspacing}{\spaceskip=\fontdimen2\font plus
\BIBentryALTinterwordstretchfactor\fontdimen3\font minus \fontdimen4\font\relax}
\providecommand{\BIBforeignlanguage}[2]{{%
\expandafter\ifx\csname l@#1\endcsname\relax
\typeout{** WARNING: IEEEtranN.bst: No hyphenation pattern has been}%
\typeout{** loaded for the language `#1'. Using the pattern for}%
\typeout{** the default language instead.}%
\else
\language=\csname l@#1\endcsname
\fi
#2}}
\providecommand{\BIBdecl}{\relax}
\BIBdecl

\bibitem[McMahan et~al.(2017)McMahan, Moore, Ramage, Hampson, and Arcas]{mcmahan2017communication}
B.~McMahan, E.~Moore, D.~Ramage, S.~Hampson, and B.~A. Arcas, ``Communication-efficient learning of deep networks from decentralized data,'' in \emph{Artificial intelligence and statistics}, 2017.

\bibitem[Diao et~al.(2021)Diao, Ding, and Tarokh]{diao2021heterofl}
E.~Diao, J.~Ding, and V.~Tarokh, ``{HeteroFL: Computation and Communication Efficient Federated Learning for Heterogeneous Clients},'' in \emph{ICLR}, 2021.

\bibitem[Zhu et~al.(2022)Zhu, Hong, Drew, and Zhou]{zhu2022resilient}
Z.~Zhu, J.~Hong, S.~Drew, and J.~Zhou, ``Resilient and communication efficient learning for heterogeneous federated systems,'' in \emph{ICML}, 2022.

\bibitem[Horvath et~al.(2021)Horvath, Laskaridis, Almeida, Leontiadis, Venieris, and Lane]{horvath2021fjord}
S.~Horvath, S.~Laskaridis, M.~Almeida, I.~Leontiadis, S.~Venieris, and N.~Lane, ``Fjord: Fair and accurate federated learning under heterogeneous targets with ordered dropout,'' \emph{NeurIPS}, vol.~34, 2021.

\bibitem[Chai et~al.(2020)Chai, Ali, Zawad, Truex, Anwar, Baracaldo, Zhou, Ludwig, Yan, and Cheng]{chai2020tifl}
Z.~Chai, A.~Ali, S.~Zawad, S.~Truex, A.~Anwar, N.~Baracaldo, Y.~Zhou, H.~Ludwig, F.~Yan, and Y.~Cheng, ``{TiFL: A tier-based Federated Learning System},'' in \emph{The 29th international symposium on high-performance parallel and distributed computing}, 2020.

\bibitem[Bao et~al.(2023)Bao, Drew, Wang, Zhou, and Niu]{bao2023fedcab}
Y.~Bao, S.~Drew, X.~Wang, J.~Zhou, and X.~Niu, ``Federated learning with client availability budgets,'' in \emph{GLOBECOM 2023 - 2023 IEEE Global Communications Conference}, 2023, pp. 1902--1907.

\bibitem[Cohen et~al.(2017)Cohen, Afshar, Tapson, and Van~Schaik]{cohen2017emnist}
G.~Cohen, S.~Afshar, J.~Tapson, and A.~Van~Schaik, ``Emnist: Extending mnist to handwritten letters,'' in \emph{2017 international joint conference on neural networks (IJCNN)}.\hskip 1em plus 0.5em minus 0.4em\relax IEEE, 2017, pp. 2921--2926.

\bibitem[Xiao et~al.(2017)Xiao, Rasul, and Vollgraf]{ribero2022fashionmnist}
H.~Xiao, K.~Rasul, and R.~Vollgraf, ``Fashion-mnist: a novel image dataset for benchmarking machine learning algorithms,'' \emph{arXiv:1708.07747}, 2017.

\bibitem[Liang and Zhao(2021)]{Liang2021NetTrajAN}
Y.~Liang and Z.~Zhao, ``Nettraj: A network-based vehicle trajectory prediction model with directional representation and spatiotemporal attention mechanisms,'' \emph{IEEE Transactions on Intelligent Transportation Systems}, vol.~23, pp. 14\,470--14\,481, 2021.

\bibitem[Forti et~al.(2020)Forti, Millefiori, Braca, and Willett]{forti20traj}
N.~Forti, L.~M. Millefiori, P.~Braca, and P.~Willett, ``Prediction oof vessel trajectories from ais data via sequence-to-sequence recurrent neural networks,'' in \emph{ICASSP 2020 - 2020 IEEE International Conference on Acoustics, Speech and Signal Processing (ICASSP)}, 2020, pp. 8936--8940.

\bibitem[Wang et~al.(2021)Wang, Jiang, Jiang, Li, and Zhao]{libcity}
J.~Wang, J.~Jiang, W.~Jiang, C.~Li, and W.~X. Zhao, ``Libcity: An open library for traffic prediction,'' in \emph{Proceedings of the 29th International Conference on Advances in Geographic Information Systems}, ser. SIGSPATIAL '21.\hskip 1em plus 0.5em minus 0.4em\relax New York, NY, USA: Association for Computing Machinery, 2021, p. 145–148.

\bibitem[Xia et~al.(2023)Xia, Li, Qi, Feng, Xu, Sun, Guo, and Jin]{xia23trajrec}
T.~Xia, Y.~Li, Y.~Qi, J.~Feng, F.~Xu, F.~Sun, D.~Guo, and D.~Jin, ``History-enhanced and uncertainty-aware trajectory recovery via attentive neural network,'' \emph{ACM Trans. Knowl. Discov. Data}, vol.~18, no.~3, dec 2023.

\bibitem[Yu et~al.(2022)Yu, Ao, Yan, Zhang, Wu, and Li]{yu22trajrec}
F.~Yu, W.~Ao, H.~Yan, G.~Zhang, W.~Wu, and Y.~Li, ``Spatio-temporal vehicle trajectory recovery on road network based on traffic camera video data,'' in \emph{Proceedings of the 28th ACM SIGKDD Conference on Knowledge Discovery and Data Mining}, ser. KDD '22.\hskip 1em plus 0.5em minus 0.4em\relax New York, NY, USA: Association for Computing Machinery, 2022, p. 4413–4421.

\bibitem[Nguyen et~al.(2018)Nguyen, Vadaine, Hajduch, Garello, and Fablet]{nguyen18recons}
D.~Nguyen, R.~Vadaine, G.~Hajduch, R.~Garello, and R.~Fablet, ``A multi-task deep learning architecture for maritime surveillance using ais data streams,'' in \emph{2018 IEEE 5th International Conference on Data Science and Advanced Analytics (DSAA)}, 2018, pp. 331--340.

\bibitem[Chen et~al.(2022)Chen, Horv{\'a}th, and Richt{\'a}rik]{chen2022optimal}
W.~Chen, S.~Horv{\'a}th, and P.~Richt{\'a}rik, ``Optimal client sampling for federated learning,'' \emph{Transactions on Machine Learning Research}, 2022.

\bibitem[Ribero et~al.(2022)Ribero, Vikalo, and De~Veciana]{ribero2022federated}
M.~Ribero, H.~Vikalo, and G.~De~Veciana, ``Federated learning under intermittent client availability and time-varying communication constraints,'' \emph{IEEE Journal of Selected Topics in Signal Processing}, 2022.

\bibitem[Jee~Cho et~al.(2022)Jee~Cho, Wang, and Joshi]{pmlr-v151-jee-cho22a}
Y.~Jee~Cho, J.~Wang, and G.~Joshi, ``Towards understanding biased client selection in federated learning,'' in \emph{Proceedings of The 25th International Conference on Artificial Intelligence and Statistics}, ser. Proceedings of Machine Learning Research, G.~Camps-Valls, F.~J.~R. Ruiz, and I.~Valera, Eds., vol. 151.\hskip 1em plus 0.5em minus 0.4em\relax PMLR, 28--30 Mar 2022, pp. 10\,351--10\,375.

\bibitem[Gu et~al.(2021)Gu, Huang, Zhang, and Huang]{gu2021fast}
X.~Gu, K.~Huang, J.~Zhang, and L.~Huang, ``Fast federated learning in the presence of arbitrary device unavailability,'' \emph{NeurIPS}, vol.~34, 2021.

\bibitem[Rodio et~al.(2023)Rodio, Faticanti, Marfoq, Neglia, and Leonardi]{rodio2023flavail}
A.~Rodio, F.~Faticanti, O.~Marfoq, G.~Neglia, and E.~Leonardi, ``{Federated Learning under Heterogeneous and Correlated Client Availability},'' in \emph{IEEE INFOCOM}, 2023.

\bibitem[Chen et~al.(2021)Chen, Yang, Saad, Yin, Poor, and Cui]{chen21wireless}
M.~Chen, Z.~Yang, W.~Saad, C.~Yin, H.~V. Poor, and S.~Cui, ``A joint learning and communications framework for federated learning over wireless networks,'' \emph{IEEE Transactions on Wireless Communications}, vol.~20, no.~1, pp. 269--283, 2021.

\bibitem[Wu et~al.(2023)Wu, Drew, and Zhou]{wu2023fedle}
J.~Wu, S.~Drew, and J.~Zhou, ``{FedLE: Federated Learning Client Selection with Lifespan Extension for Edge IoT Networks},'' in \emph{IEEE ICC}, 2023.

\bibitem[Arouj and Abdelmoniem(2022)]{arouj22battery}
A.~Arouj and A.~M. Abdelmoniem, ``Towards energy-aware federated learning on battery-powered clients,'' in \emph{Proceedings of the 1st ACM Workshop on Data Privacy and Federated Learning Technologies for Mobile Edge Network}, ser. FedEdge '22.\hskip 1em plus 0.5em minus 0.4em\relax New York, NY, USA: Association for Computing Machinery, 2022, p. 7–12.

\bibitem[Yemini et~al.(2022)Yemini, Saha, Ozfatura, G\"{u}nd\"{u}z, and Goldsmith]{yemini22semi}
M.~Yemini, R.~Saha, E.~Ozfatura, D.~G\"{u}nd\"{u}z, and A.~J. Goldsmith, ``Semi-decentralized federated learning with collaborative relaying,'' in \emph{2022 IEEE International Symposium on Information Theory (ISIT)}.\hskip 1em plus 0.5em minus 0.4em\relax IEEE Press, 2022, p. 1471–1476.

\bibitem[Liu et~al.(2022)Liu, Zeng, Zhang, Zhou, Mu, Zhang, Zhang, and Zhu]{liu22fedtad}
F.~Liu, C.~Zeng, L.~Zhang, Y.~Zhou, Q.~Mu, Y.~Zhang, L.~Zhang, and C.~Zhu, ``Fedtadbench: Federated time-series anomaly detection benchmark,'' in \emph{2022 IEEE 24th Int Conf on High Performance Computing and Communications; 8th Int Conf on Data Science and Systems; 20th Int Conf on Smart City; 8th Int Conf on Dependability in Sensor, Cloud and Big Data Systems and Application (HPCC/DSS/SmartCity/DependSys)}.\hskip 1em plus 0.5em minus 0.4em\relax Los Alamitos, CA, USA: IEEE Computer Society, dec 2022, pp. 303--310.

\bibitem[Liu et~al.(2021)Liu, Zhu, Gao, and Xu]{liu21forcast}
Z.~Liu, Z.~Zhu, J.~Gao, and C.~Xu, ``Forecast methods for time series data: A survey,'' \emph{IEEE Access}, vol.~9, pp. 91\,896--91\,912, 2021.

\bibitem[Fang et~al.(2022)Fang, Zhu, Zhang, Yu, and Xue]{Fang2022HeterogeneousTF}
J.~Fang, C.~H. Zhu, P.~Zhang, H.~Yu, and J.~Xue, ``Heterogeneous trajectory forecasting via risk and scene graph learning,'' \emph{IEEE Transactions on Intelligent Transportation Systems}, vol.~24, pp. 12\,078--12\,091, 2022.

\bibitem[Li et~al.(2020)Li, Sahu, Zaheer, Sanjabi, Talwalkar, and Smith]{tian2022fedprox}
T.~Li, A.~K. Sahu, M.~Zaheer, M.~Sanjabi, A.~Talwalkar, and V.~Smith, ``{Federated Optimization in Heterogeneous Networks},'' in \emph{Proceedings of Machine Learning and Systems (MLSys)}, 2020.

\bibitem[Li et~al.(2021)Li, He, and Song]{qinbin2022moon}
Q.~Li, B.~He, and D.~Song, ``{Model-Contrastive Federated Learning},'' in \emph{IEEE/CVF CVPR}, 2021.

\bibitem[An-Yuhang-ace(2021)]{vesseldata}
An-Yuhang-ace. (2021) Vessel trajectory prediction.

\bibitem[Zheng(2011)]{zheng2011t-drive}
Y.~Zheng, ``T-drive trajectory data sample,'' August 2011, t-Drive sample dataset.

\end{thebibliography}

\end{document}